\documentclass{article}

     \usepackage[nonatbib, final]{neurips_2020}
     
\usepackage{cite}

\usepackage[utf8]{inputenc} % allow utf-8 input
\usepackage[T1]{fontenc}    % use 8-bit T1 fonts
\usepackage{hyperref}       % hyperlinks
\usepackage{url}            % simple URL typesetting
\usepackage{booktabs}       % professional-quality tables
\usepackage{amsfonts}       % blackboard math symbols
\usepackage{nicefrac}       % compact symbols for 1/2, etc.
\usepackage{microtype}      % microtypography
\usepackage{multirow}
\usepackage{graphicx}
\usepackage{subcaption}
\usepackage{amsmath}
\usepackage{mathtools}  
\usepackage{hyperref}
\usepackage{hhline}
\usepackage{wrapfig}

\usepackage{bm}

\title{NanoFlow: Scalable Normalizing Flows with Sublinear Parameter Complexity}

\author{%
  Sang-gil Lee \hspace{40pt} Sungwon Kim \hspace{40pt} Sungroh Yoon\thanks{Corresponding author} \\
  Data Science \& AI Lab. \\
  Seoul National University \\
  \texttt{\{tkdrlf9202, ksw0306, sryoon\}@snu.ac.kr} \\
}

\begin{document}

\maketitle

\begin{abstract}
  Normalizing flows (NFs) have become a prominent method for deep generative models that allow for an analytic probability density estimation and efficient synthesis. However, a flow-based network is considered to be inefficient in parameter complexity because of reduced expressiveness of bijective mapping, which renders the models unfeasibly expensive in terms of parameters. We present an alternative parameterization scheme called NanoFlow, which uses a single neural density estimator to model multiple transformation stages. Hence, we propose an efficient parameter decomposition method and the concept of flow indication embedding, which are key missing components that enable density estimation from a single neural network. Experiments performed on audio and image models confirm that our method provides a new parameter-efficient solution for scalable NFs with significant sublinear parameter complexity.
\end{abstract}

\section{Introduction}
Flow-based models have become a prominent approach for density estimation and generative models in recent times. These models are based on normalizing flows (NFs) \cite{rezende2015variational}, wherein a deep invertible model is trained with an analytically estimated likelihood of data. The model learns a bijective mapping between the data and a known prior (typically isotropic Gaussian), and its reverse operation synthesizes realistic samples from the prior. Compared with the variational autoencoder \cite{kingma2013auto} and generative adversarial network \cite{goodfellow2014generative}, NFs exhibit the distinct characteristic of an exact probability density estimation from a principled maximum likelihood training. When combined with non-autoregressive coupling methods \cite{dinh2016density, kingma2018glow}, NFs become a powerful generative model that offers a significantly simplified training and efficient inference.

Since the introduction of the framework into neural networks, the existing studies on flow-based models have focused on improving the expressiveness of the bijective operation \cite{ho2019flow++, kingma2018glow, durkan2019neural, chen2019residual}. However, parameter complexity and memory efficiency are less emphasized by the research community. The small efforts to maximize the expressiveness \textit{under a specified amount of capacity of the neural network} has recently become problematic when expanding a flow-based model for real-world applications. A notable example is the waveform synthesis model \cite{prenger2019waveglow, kim2019flowavenet}. Although the aforementioned studies have achieved audio generation faster than real-time (thereby removing the slow inference bottleneck of WaveNet \cite{van2016wavenet}), they resulted in an increase in the number of parameters by an order of magnitude, which is unfeasibly expensive in terms of memory. Hence, building a complex, scalable, and \textit{memory-efficient} flow-based model remains challenging.

This scenario raised a question: \textit{Is it true that NFs require a significantly larger network capacity to perform expressive bijections, or is the representational power of deep neural networks inefficiently utilized?} We argue that studies regarding NFs should consider the parameter complexity, where the expressiveness of multiple flows is not necessarily accompanied by a linearly growing number of parameters.

In this study, we challenge the typical assumption in building flow-based models and aim to decouple the required number of parameters and the expressiveness of multiple bijective operations for flow-based models. We present NanoFlow, an alternative parameterization scheme for NFs that operates on a single neural density estimator. Because the shared density estimator is applied to multiple stacks of flows, the parameter requirement is no longer proportional to the number of flows, and the memory footprint is significantly reduced. Consequently, NanoFlow can consistently improve its expressiveness by stacking flows without sacrificing parameter efficiency.

Our results indicated that using a conventional notion of weight sharing did not yield a good performance on flow-based models, which nullifies the potential benefits. To achieve the concept of a shared neural density estimator, we demonstrate several parameter-efficient solutions for increasing the flexibility of NanoFlow. We show that decomposing a deep hidden representation estimated by the shared neural network and the projected densities from the representation can significantly enhance the expressiveness of NanoFlow with the addition of a few parameters. Furthermore, we also demonstrate that conditioning the shared estimator with our flow indication embedding can remedy the modeling difficulties of multiple densities from a single estimator without dissatisfying any invertibility constraints.

Additionally, we provide a deeper analysis of the condition under which our method yields the a higher number of benefits. Specifically, we assess the effectiveness of the single density estimator by varying the amount of autoregressive structural bias into the model. Our results demonstrate that our method performs best on bipartite flows, which provides an expanded narrative on our belief regarding the performance gap between non-autoregressive and autoregressive models. In summary, our study is the first to focus on a systematic assessment for enabling scalable NFs with an almost constant parameter complexity.

\section{Background}
NFs learn the bijective mapping between data and a known prior. The prior is typically constructed as an isotropic Gaussian, and the reverse of the bijective mapping can synthesize the data from the noise sampled from the prior. Formally, NFs learn the bijective function $f(\bm{x})=\bm{z}$, which transforms a complex data probability distribution $P_{\bm{X}}$ into a simple known prior $P_{\bm{Z}}$ with the same dimension. We can analytically compute the probability density of real data $\bm{x}$ using the change of variables formula:
\begin{equation}
\label{eq1}
\log P_{\bm{X}}(\bm{x}) = \log P_{\bm{Z}}(\bm{z}) + \log|\det(\frac{\partial f(\bm{x})}{\partial \bm{x}})|,
\end{equation}
where $\det(\frac{\partial f(\bm{x})}{\partial \bm{x}})$ is a Jacobian determinant of the function $f(\bm{x})=\bm{z}$. To enhance the expressiveness of $f$, NF models decompose the function into multiple flows as follows:
\begin{equation}
\label{eq4}
f = f^{K} \circ f^{K-1} \circ ... \circ f^{1}(\bm{x}),
\end{equation}
where $K$ is the number of flows defined by the model. Using the notations $\bm{x}$ = $\bm{z}^0$ and $\bm{z}=\bm{z}^K$, each $f^k(\bm{z}^{k-1})=\bm{z}^k$ learns the intermediate densities between $\bm{x}$ and $\bm{z}$, and  $\log P_{\bm{X}}$ can be re-expressed as follows:
\begin{equation}
\label{eq_nf}
\log P_{\bm{X}}(\bm{x}) = \log P_{\bm{Z}}(\bm{z}) + \sum_{k=1}^{K} \log|\det(\frac{\partial {f^{k}}(\bm{z}^{k-1})}{\partial \bm{z}^{k-1}})|.
\end{equation}
Because the determinant typically requires $O(n^3)$ computing time (where $n$ is the dimension of the data), NF models are designed to maintain a triangular Jacobian \cite{dinh2014nice, kingma2016improved, papamakarios2017masked}. By maintaining a triangular Jacobian, the determinant becomes easy to compute, and the model becomes computationally tractable for both forward and inverse functions.

Our mathematical notation for the coupling transformation follows that of the WaveFlow \cite{ping2020waveflow}. Although the study focused on waveform synthesis, it provides a unified view from bipartite to autoregressive flows, which subsumes a wide range of flow-based models. We note that \cite{papamakarios2017masked} also provides a relevant analysis regarding the relationship between autoregressive and bipartite flows.

Formally, for training data $\bm{x}$, assume that we split $\bm{x}$ into $G$ groups as $\{ \bm{X}_1, ..., \bm{X}_G\}$. The model is trained to learn the bijective mapping between $\bm{X}$ and a prior $\bm{Z}$ with the same dimension. This is achieved by applying an affine transformation $f: \bm{X} \rightarrow \bm{Z}$ which models a sequential dependency between the grouped data as follows:
\begin{equation}
\label{eq2}
\bm{Z}_{i} = \sigma_{i}(\bm{X}_{<i}; \theta) \cdot \bm{X}_{i} + \mu_{i}(\bm{X}_{<i}; \theta), \quad i=1, ..., G,
\end{equation}
where $\bm{X}_{<i}$ refers to all the partitions of the data before the $i$-th group, $\bm{X}_i$; $\sigma$ and $\mu$ are the scale and shift variables estimated by the neural networks, respectively. From the sampled noise $\bm{Z}$, the inverse of the affine transformation $f^{-1}: \bm{Z} \rightarrow \bm{X}$ sequentially generates $\bm{X}$ as follows:
\begin{equation}
\label{eq3}
\bm{X}_{i} = \frac{\bm{Z}_{i} - \mu_{i}(\bm{X}_{<i}; \theta)}{\sigma_{i}(\bm{X}_{<i}; \theta)}, \quad i=1, 2, ..., G.
\end{equation}
The model becomes a purely autoregressive flow when $G=dim(\bm{x})$. Conversely, the equations theoretically represent bipartite flows when $G=2$. Increasing the number of groups introduces a higher amount of autoregressive structural bias into the model, at a cost of $O(G)$ inference latency.

As previously mentioned, the entire bijective function $f: \bm{X} \rightarrow \bm{Z}$ is decomposed into $K$ flows as $f = f^{K} \circ f^{K-1} \circ ... \circ f^{1}(\bm{X})$, where we use the notation $\bm{X}=\bm{Z}^{0}$ and $\bm{Z}=\bm{Z}^{K}$. Each $f^{k}: \bm{Z}^{k-1} \rightarrow \bm{Z}^{k}$ is parameterized by separate neural networks $\theta^{k}$, whereas each $\theta^{k}$ estimates the intermediate density of $\bm{Z}^{k}$ by computing $\sigma^{k}$ and $\mu^{k}$ for the flow operation. For clarity, we consider the notation of $\bm{X}$ as the input and $\bm{Z}$ as the output for $f^{k}$. We re-write $f^{k}$ for completeness as follows:
\begin{equation}
\label{eq5}
\bm{Z}_{i} = \sigma_{i}(\bm{X}_{<i}; \theta^{k}) \cdot \bm{X}_{i} + \mu_{i}(\bm{X}_{<i}; \theta^{k}).
\end{equation}

The above formula describes the affine transformation as a bijective coupling operation. Various other classes of coupling exist in the literature \cite{durkan2019neural, ho2019flow++}.

\section{NanoFlow}
\label{nanoflow}
\begin{figure*}[t]
    \centering
    \begin{subfigure}[t]{.25\linewidth}
        \centering
        \includegraphics[height=4.8cm]{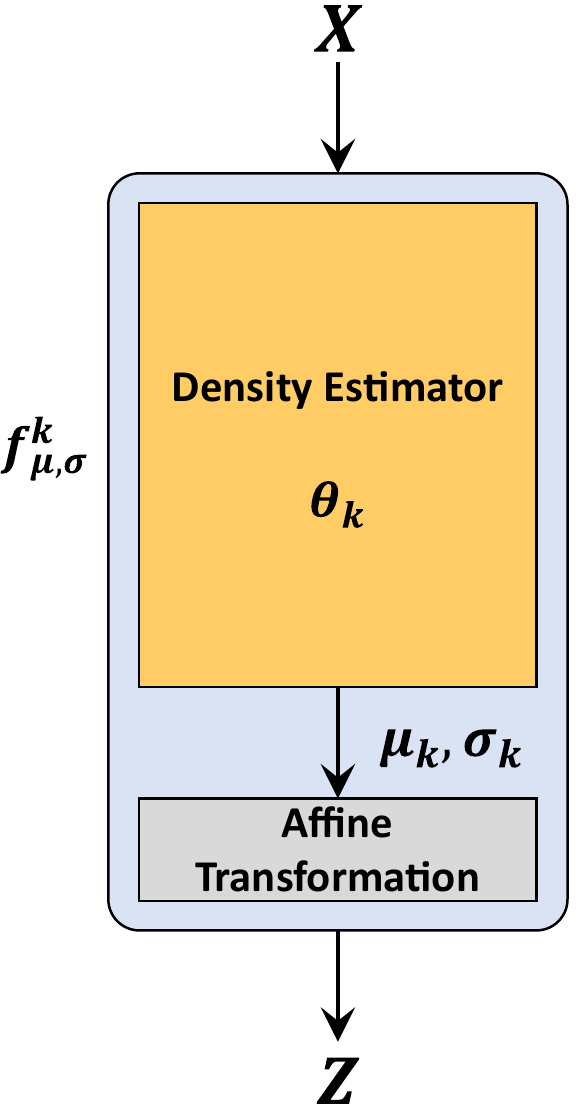}
        \caption{Baseline}
        \label{fig1:figure_a}
    %\end{subfigure}\hspace*{\fill}
    \end{subfigure}%
    \hspace{1em}
    \begin{subfigure}[t]{.25\linewidth}
        \centering
        \includegraphics[height=4.8cm]{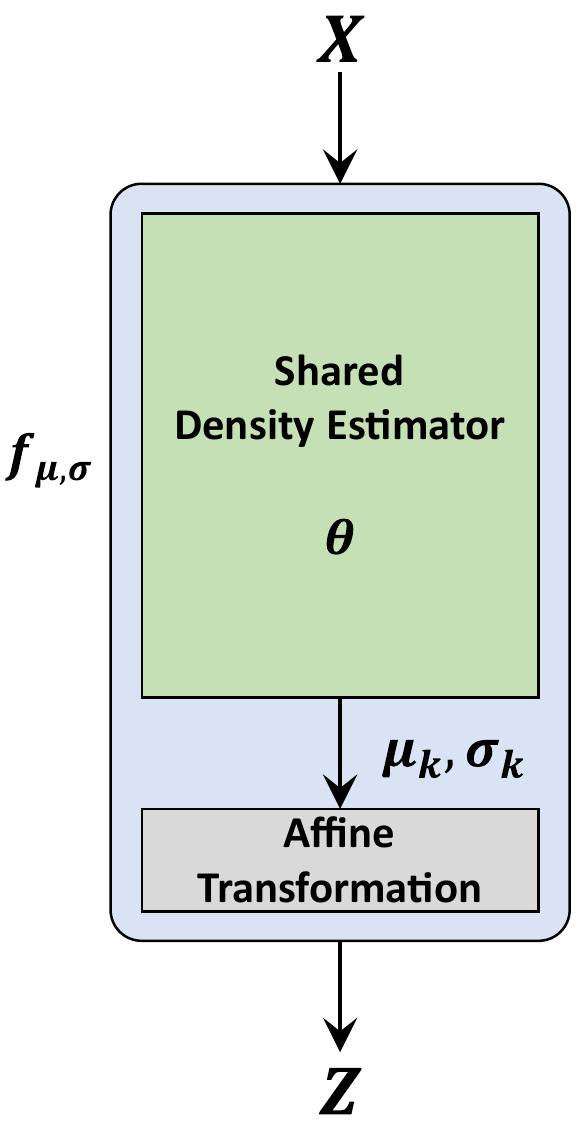}
        \caption{NanoFlow-naive}
        \label{fig1:figure_b}
    %\end{subfigure}\hspace*{\fill}
    \end{subfigure}%
    \hspace{1em}
    \begin{subfigure}[t]{.4\linewidth}
        \centering
        \includegraphics[height=4.8cm]{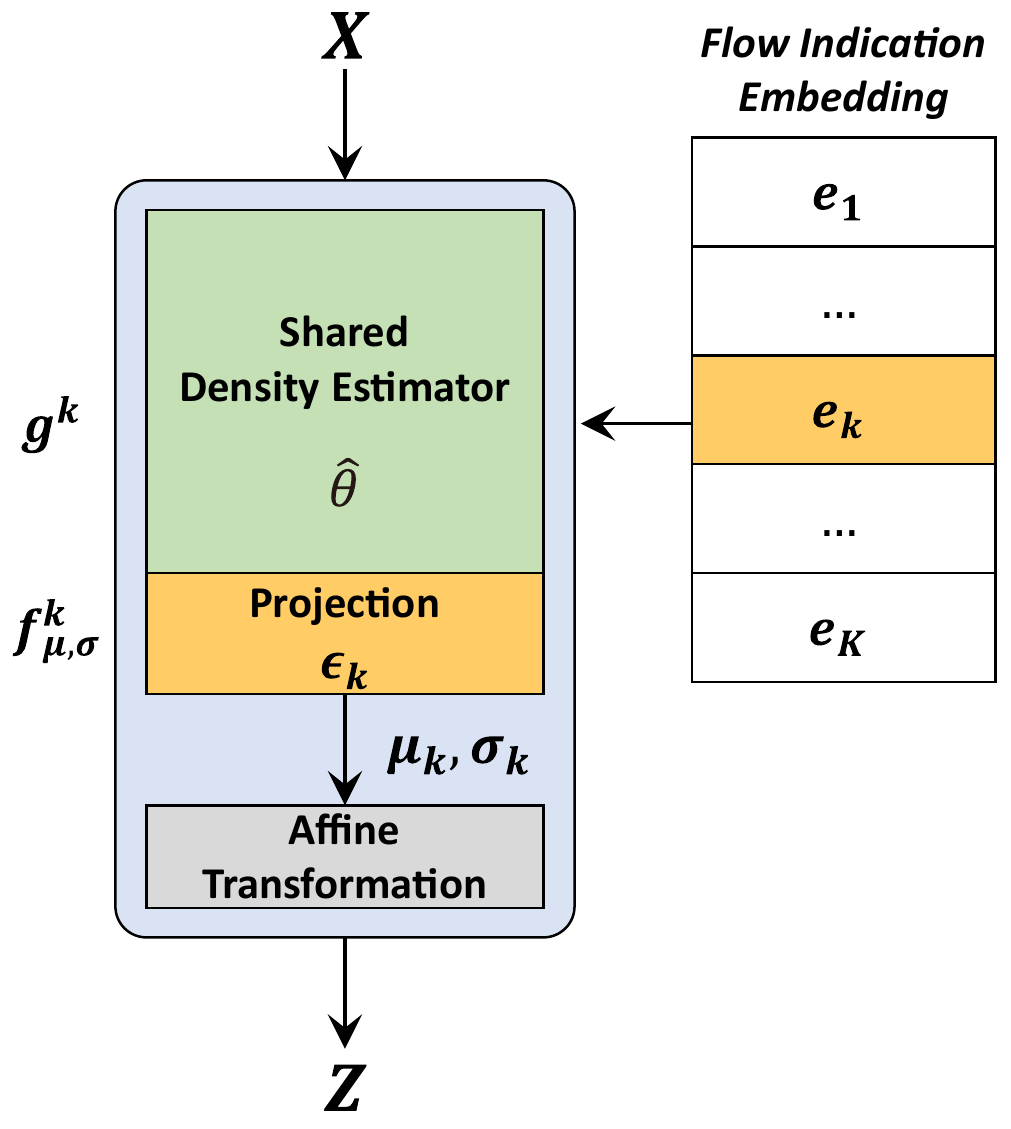}
        \caption{NanoFlow}
        \label{fig1:figure_c}
    \end{subfigure}
    \caption{High-level overview of NanoFlow. (a) Conventional NFs employ separate neural networks as a density estimator for each flow. (b) NanoFlow-naive shares an entire part of the neural network for density estimation with multiple flow steps. (c) NanoFlow decomposes the estimator into two parts—one for the deep shared latent space representation augmented by flow indication embedding, and another for separate projection layers employed to each flow.}
    \label{fig1}
\end{figure*}

In this section, we present NanoFlow, a new alternative parameterization scheme for a flow-based model (Figure \ref{fig1}). The main goal of NanoFlow is to decouple the expressiveness of the bijections and the parameter efficiency of density estimation from neural networks. We initially describe a core change in the design of the neural architecture by decomposing the parameters for neural density estimation and sharing parameters across flows.

\subsection{Parameter sharing and decomposition}
We reformulated $f_{\mu, \sigma}^{k}$ using a single shared neural network $f_{\mu, \sigma}$, parameterized by $\theta$. Based on this framework, all $\sigma^{k}$ and $\mu^{k}$ for each flow were estimated by the shared $f_{\mu, \sigma}$. This formulation can reduce the number of parameters by a fraction of the number of flows by $\frac{1}{K}$, and we call this variant, the NanoFlow-naive. However, as our experimental results suggest, this aggressive re-use of parameters is unsuitable for modeling multiple densities that suffer from severe degradation in performance, as it completely nullifies the potential benefit of the $O(1)$ memory footprint.

Based on these observations, we propose to relax the constraint of the shared estimator by decomposing the shared model into a network that computes a hidden representation and a projection layer that estimates the densities. The function is decomposed into $f_{\mu, \sigma}^{k} \circ g$, where $g(\cdot;\hat{\theta})$ is the shared estimator parameterized by $\hat{\theta}$, excluding the projection layer, that is, each $f_{\mu, \sigma}^{k}$ has separate parameters for the projected intermediate density by computing $\sigma^{k}$ and $\mu^{k}$.

Assuming that $g$ has sufficient capacity for density estimation, the projection layer can be as shallow as a $1\times1$ convolution; hence, the number of parameters is negligible in comparison with $\hat{\theta}$. Using this decomposition, we can construct NanoFlow with an arbitrary number of flows. Interestingly, this alternative scheme was pivotal for achieving the competitive performance of NanoFlow. We observed that the likelihood of the data continuously increased as we stacked additional flows without sacrificing the efficiency of weight sharing. Hence, we can re-write $f^{k}$ as follows:
\begin{equation}
\label{eq2}
\bm{Z}_{i} = \sigma_{i}(\bm{X}_{<i}; \hat{\theta}, \epsilon^{k}) \cdot \bm{X}_{i} + \mu_{i}(\bm{X}_{<i}; \hat{\theta}, \epsilon^{k}),
\end{equation}
where $\epsilon^{k}$ is the parameter of the separate projection layer assigned to each flow.

\begin{table}
  \caption{Comparison of parameterization scheme of $f^k$ between methods for bijection. $K$ is the total number of flows defined by the model, and $|\bullet|$ is the number of parameters of the neural network with the designated letters.}
  \label{complexity_analysis}
  \centering
  \begin{tabular}{lcccc}
    \toprule
    Method & $f^{k}: \bm{X}_{i} \rightarrow \bm{Z}_{i} = \sigma_{i}^{k} \cdot \bm{X}_{i} + \mu_{i}^{k}, i=1, ..., G$ & Parameters                   \\
    \midrule
    WaveFlow (baseline)    & $ \mu_{i}^{k}, \sigma_{i}^{k} = f_{\mu, \sigma}^{k}(\bm{X}_{<i}; \theta^k) $ & $\sum_{k=1}^{K} |\theta^k|$\\
    \addlinespace[0.05cm]
    NanoFlow-naive & $ \mu_{i}^{k}, \sigma_{i}^{k} = f_{\mu, \sigma}(\bm{X}_{<i}; \theta) $ & $|\theta|$\\
    \addlinespace[0.05cm]
    NanoFlow-decomp    & $ \mu_{i}^{k}, \sigma_{i}^{k} = f_{\mu, \sigma}^{k}(g(\bm{X}_{<i}; \hat{\theta}); \epsilon^k) $ & $|\hat{\theta}| + \sum_{k=1}^{K} |\epsilon^k|$\\
    \addlinespace[0.05cm]
    NanoFlow (proposed)    & $ \mu_{i}^{k}, \sigma_{i}^{k} = f_{\mu, \sigma}^{k}(g^{k}(\bm{X}_{<i}; \hat{\theta}, \bm{e}^k); \epsilon^k) $ & $|\hat{\theta}| + \sum_{k=1}^{K} (|\epsilon^k |+ |\bm{e}^k|)$ \\
    \bottomrule
  \end{tabular}
\end{table}

\subsection{Flow indication embedding}
Even when the parameter decomposition described above is used, the shared estimator $g$ must learn multiple intermediate densities of bijective transformations without context. Hence, we introduce a key missing module, which we name flow indication embedding, to enable the shared model to simultaneously learn multiple bijective transformations. Because the flow-based model is based on the bijective function, the embedding must be available \textit{a priori} for application to the reverse operation.

For each $f^{k}$, we define an embedding vector $\bm{e}^{k}\in \mathbb{R}^{D}$, where $D$ is the dimension of the embedding. Subsequently, we feed the embedding to the shared model $g(\cdot;\hat{\theta})$ as an additional context. From the embedding $\bm{e}^{k}$, we can further guide $g(\cdot;\hat{\theta})$ to learn multiple intermediate densities with minimal addition of parameters, transforming it into $g^{k}(\cdot;\hat{\theta}, \bm{e}^{k})$. Because the order of flow operations is pre-defined, we can use $\bm{e}^{k}$ by feeding it to the shared estimator in the reverse order during inference, that is, our embedding does not dissatisfy any invertibility constraints.

The optimal injection of the embedding into $g^{k}$ may depend on the neural architecture. We investigated three candidates: 1. concatenative embedding, in which we augment the input with the embedding vector at the start of each flow;  2. additive bias, in which for each layer inside $g^{k}$, $\bm{e}^k$ provides a channel-wise bias projected from additionally defined $1\times1$ convolutional layers; 3. multiplicative gating, in which we employ independent per-channel scalars inside $g^{k}$ for each flow that controls the propagation of the convolutional feature map according to the specified flow steps. Note that the aforementioned methods involve a negligible number of additional parameters, do not dissatisfy the invertibility, and impose a minimal effect on the inference latency. See Appendix \ref{appendix:a} for a more detailed description.

Table \ref{complexity_analysis} summarizes the parameterization scheme of $f^k$ and its complexity. The parameter efficiency of NanoFlow is due to employing a single neural density estimator, $g$, for multiple flow operations. NanoFlow-naive incorporates a conventional weight sharing and NanoFlow-decomp relaxes the constraint of intermediate density estimations by employing separate $\epsilon^k$ for each flow. The final proposed model, NanoFlow, further increases the parameter efficiency of $g^{k}$ by incorporating the flow indication embedding $\bm{e}^k$. We emphasize that $\hat{\theta}$ embodies the majority of the parameters from the model.

\section{Experiments}
In this section, we present a systematic assessment of the effectiveness of NanoFlow. We initially present the experimental results from an audio generative model with WaveFlow \cite{ping2020waveflow} as the baseline architecture, combined with an extensive ablation study. Next, we provide a likelihood ratio analysis of NanoFlow by varying the amount of autoregressive structural bias into both models, which evaluates the conditions under which NanoFlow yields more benefits. Finally, we investigate the generalizability of our methods by performing density modeling on the image domain, with Glow \cite{kingma2018glow} as the reference model.

\begin{table}
  \caption{Model comparison. We report the number of model parameters in millions (M), a log-likelihood (LL) on the test set, a subjective five-scale mean opinion score (MOS) on naturalness with 95 \% confidence interval, and a synthesis speed using a single Nvidia V100 GPU with half-precision arithmetic. MOS on ground-truth audio is 4.58 $\pm$ 0.06.}
  \label{mos}
  \centering
  \begin{tabular}{lccccc}
    \toprule
    %\multicolumn{2}{c}{Part}                   \\
    Method     & Channels & Parameters (M)     & LL & MOS & kHz\\
    \midrule
    WaveFlow (Re-impl) & 128 & 22.336 & 5.2059 &  4.11 $\pm$ 0.08 & 347 \\
    WaveFlow (Re-impl) & 64 & 5.925  & 5.1357 &  3.52 $\pm$ 0.09  & 828 \\
    NanoFlow-naive     & 128 & 2.792 & 5.1247 &    3.23 $\pm$ 0.09 & 376 \\
    NanoFlow     & 128 & 2.819      & 5.1586 & 3.63 $\pm$ 0.09 & 362 \\
    NanoFlow (K=16)  & 128 & 2.845   & 5.1873 & 3.82 $\pm$ 0.08 & 186 \\
    \bottomrule
  \end{tabular}
\end{table}

\subsection{Audio generation results}
\label{audio_generation_results}
For the performance evaluation of waveform generation, we used the LJ speech dataset \cite{ljspeech17}, which is a 24-h single-speaker speech dataset containing 13,100 audio clips. We used the first 10$\%$ of the audio clips as the test set and the remaining 90$\%$ as the training set. We used the audio preprocessing and mel-spectrogram construction pipeline provided by the official WaveGlow implementation \cite{prenger2019waveglow}. Specifically, we used an 80-band log-scale mel-spectrogram condition with an FFT size of 1,024, a hop size of 256, and a window size of 1,024. We used a maximum frequency of 8,000Hz for the STFT without audio volume normalization, and we set the noise distribution to $\bm{Z}_{i} \sim \mathcal{N}(0, 1)$, which is the default setting from the open-source WaveGlow.

We used the default architecture described in \cite{ping2020waveflow} with $G=16$ for WaveFlow and NanoFlow. We constructed the models with eight flows unless otherwise specified and used the permutation strategy of reversing the order of the group dimensions per flow for both models. Our selection of flow indication embedding is a combination of additive bias and multiplicative gating, as WaveNet-based \cite{van2016wavenet} architecture features a natural method of utilizing additive bias as global conditioning augmented by a gated residual path \cite{van2016pixel}. We used $D=512$ for $\bm{e}^{k}\in \mathbb{R}^{D}$ in the default eight-flows model and $D=1024$ in the 16-flows variant.

We trained all models for 1.2 M iterations with a batch size of eight and an audio clip size of 16,000, using an Nvidia V100 GPU. We used the Adam optimizer \cite{kingma2014adam} with an initial learning rate of $10^{-3}$, and we annealed the learning rate by half for every 200 K iterations. For the evaluation, we applied checkpoint averaging over 10 checkpoints with 5 K iteration intervals. We sampled the audio at a temperature of 1.0.

Table \ref{mos} shows an objective performance measure of log-likelihood (LL) on the test set as well as a subjective and relative audio quality evaluation with a five-scale mean opinion score (MOS) on naturalness using the Amazon Mechanical Turk. Furthermore, we provide the audio synthesis speed in kilohertz using a single Nvidia V100 GPU with half-precision arithmetic.

The results show that our method can synthesize waveforms with a slight quality degradation against the baseline while only using approximately $1/8$ of the parameters. However, the NanoFlow-naive failed to perform competitively even against a 64-channel variant of WaveFlow. This suggests that for flow-based models, a strict constraint of $O(1)$ memory requirement severely degenerates the modeling capability. NanoFlow-decomp performed slightly better than NanoFlow-naive with a likelihood score of 5.13, which was still insufficient as a competitive alternative.

On the contrary, NanoFlow provided significantly enhanced expressiveness, with a negligible number of additional parameters from the decomposition technique with flow indication embedding. By stacking double the steps of flows, we further verified that the enhanced expressiveness of the flows was no longer proportional to the capacity of the deep generative model. Consistent with the results from a previous work \cite{ping2020waveflow}, we observed that the subjective MOS scores exhibited good alignment with the objective likelihood scores.

\begin{table}
  \caption{LL ratio results with varying amount of autoregressive structural bias on the number of groups. Lower values indicate higher similarity in probability density modeling performance between the two models.}
  \label{LLR}
  \centering
  \begin{tabular}{lcccccc}
    \toprule
    %\multicolumn{2}{c}{Part}                   \\
    \multicolumn{2}{c}{Number of Groups (G)}&4 & 8 & 16 & 32 & 64\\
    \midrule
    \multirow{2}{*}{LL} & WaveFlow (5.96 M) & 4.9785 & 5.0564 & 5.1241 & 5.141 & 5.1586 \\
    & NanoFlow (0.75 M) & 4.9513 & 5.0271 & 5.0797 & 5.0927 & 5.111  \\
    \midrule
    \multicolumn{2}{c}{LL ratio} &\textbf{0.0272} & 0.0293 & 0.0444 & 0.0483 & 0.0476\\
    \bottomrule
  \end{tabular}
\end{table}

\subsection{Likelihood ratio analysis with autoregressive structural bias}
Our reference model, WaveFlow \cite{ping2020waveflow}, provided a unified view of the expressiveness of flow-based models by incorporating a fixed amount of autoregressive structural bias into the architecture. The model provides a hybrid method in which the autoregressive bias is proportional to the number of group dimensions. In this section, we provide an expanded narrative on the performance gap between the non-autoregressive and autoregressive flows by adjusting the amount of bias for both WaveFlow and NanoFlow. We trained each model with 64 channels for 500 K iterations with a batch size of two for varying degrees of the group dimension. We used $D=128$ for the NanoFlow embedding.

Table \ref{LLR} quantitatively shows the expressiveness of autoregressive bias. As we enforce a higher amount of the autoregressive structure into the model, we can achieve a more expressive model under the same network capacity. However, this is at the expense of sequential inference, which has been reported in previous studies \cite{kingma2018glow, van2016wavenet, oord2018parallel}.

In addition, we provide the LL ratio between WaveFlow and NanoFlow, where we measure the gap in modeling capability by introducing a shared neural density estimator. Most importantly, we observed a nearly monotonic decrease in the performance gap of NanoFlow as we decreased the number of groups. This further provides an insight into our effectiveness in utilizing the capacity of the deep generative network. If we impose a lower amount of the explicit dependency between partitions of data, we can extract a deep shared latent representation that is easier to manipulate by our flow indication embedding. In other words, we can expect a wide range of flow-based models with bipartite coupling to benefit significantly from the parameterization scheme of NanoFlow.

\begin{table}
  \caption{Unconditional image density estimation results with bits per dimension (bpd) on CIFAR10 under uniform dequantization. Results with $\dagger$ were taken from the existing literature \cite{ finlay2020train}.}
  \label{ablation}
  \centering
  \begin{tabular}{lcccc} \toprule
Method && Parameters (M) && bpd \\ \midrule
Glow (256 channels) && 15.973 && 3.40 \\
Glow (512 channels) \cite{kingma2018glow} $\dagger$ && 44.235 && 3.35 \\
Glow-large && 287.489 && 3.30 \\
RQ-NSF (C) \cite{durkan2019neural} $\dagger$ && 11.8 && 3.38 \\
FFJORD \cite{grathwohl2019ffjord} $\dagger$ & &1.359 && 3.40 \\
MintNet \cite{song2019mintnet} $\dagger$ && 27.461 && 3.32 \\
Flow++ \cite{ho2019flow++} $\dagger$ && 32.3 && 3.28 \\ 
ResidualFlow \cite{chen2019residual} $\dagger$ && 25.174 && 3.28 \\ \midrule
NanoFlow-naive && 9.263 && 3.40 \\
NanoFlow-decomp && 9.935 && 3.32 \\
NanoFlow && 10.113 && \textbf{3.27} \\
NanoFlow (K=48) && 10.718 && \textbf{3.25} \\
\bottomrule
\end{tabular}
\end{table}

\subsection{Image density modeling results}
\label{image_density_modeling}

To demonstrate that our method is applicable to any configuration of NF and data domains, we assessed the effectiveness of NanoFlow's parameterization scheme to Glow \cite{kingma2018glow}. We used the training configurations of an open-source implementation as described in \cite{kingma2018glow}. We trained Glow, NanoFlow, and its ablations on the CIFAR10 dataset for 3,000 epochs, where all model configurations reached saturation in performance. We used 256 channels and a batch size of 64 for all configurations for an extensive ablation study under a fixed computational budget.

Because NanoFlow is designed to leverage the shared density estimator with sufficient capacity, we increased the number of convolutional layers to six, and modified the kernel size to $3\times3$ for all layers. We changed the kernel size of the separate density projection layers to $1\times1$ to maintain the nearly constant memory footprint of NanoFlow. We refer to the model with this modified architecture without the application of our method as Glow-large. This model serves as an upper bound on modeling performance, but the parameter complexity is increased. We trained the original model with the exact network topology from \cite{kingma2018glow} together with Glow-large to completely assess the capability of NanoFlow. Because Glow uses a multi-scale architecture \cite{dinh2016density}, NanoFlow is applied by sharing the estimator separately for each scale. We used concatenative embedding together with multiplicative gating as the flow indication embedding. For $\bm{e}^{k}\in \mathbb{R}^{D}$, we used $D=64$ for the default 32 flows per scale, and $D=192$ for a scaled-up model with 48 flows per scale.

As presented in Table \ref{ablation}, we observed that the reference Glow model scaled with a higher network capacity, at the cost of the increased parameters and decreased return. NanoFlow-naive failed to perform competitively, even with the increased capacity of the shared estimator. This suggests that even if a more powerful neural network is introduced, a critical bottleneck exists when modeling multiple flows from the single model without applying our method.

Unlike the waveform synthesis results, applying only the decomposition technique was sufficient to outperform NanoFlow-naive by a large margin. The performance was further improved using flow indication embedding. NanoFlow with the default number of flows (32 steps per scale) exhibited better performance than Glow-large, which has more than 28 times more parameters. This illustrates that in NFs, leveraging the shared neural network would be easier to train and more scalable with better inductive bias, provided with proper methods as shown by NanoFlow, than employing separate estimators where each neural network should learn the intermediate probability densities from scratch.

When we scaled up the model to 48 flows per scale, we observed an additional gain in performance from the shared estimator, further confirming the scalability of the proposed method. NanoFlow was able to achieve competitive performance compared to studies with more complex non-affine coupling \cite{ho2019flow++, song2019mintnet, durkan2019neural, chen2019residual}, indicating potential benefits of deep and shared latent representation. Overall, the density estimation results with bits per dimension were consistent with the audio generation results. The effectiveness of our method was further highlighted in this setup with bipartite coupling, which further confirms our findings from the likelihood ratio analysis in the preceding section. See Appendix \ref{appendix:b} for the additional results and Appendix \ref{appendix:c} for the sampled images from the models.

\section{Related Work}
\label{rel_works}

\subsection{Improving coupling transformations}
Since the introduction of NFs into neural networks \cite{dinh2014nice, rezende2015variational}, most studies have focused on composing a flexible bijection for better expressiveness \cite{papamakarios2017masked, dinh2016density, kingma2018glow, ho2019flow++, durkan2019neural, hoogeboom2019emerging, ma2019macow}. Building a more complex bijection can also achieve better memory efficiency by attaining the desired level of complexity under fewer flow operations. Our study provides an orthogonal perspective on this topic with a specific focus on the parameterization of a scalable NF \textit{under a specified network capacity}, where we systematically assess the feasibility of employing a single shared neural density estimator for multiple flow steps. Because our parameterization scheme is agnostic to any setup of flow-based models and coupling operator, we can apply any off-shelf bijections into our framework, together with improved methods for training NFs \cite{ho2019flow++}.

\subsection{Parameter sharing}
 The concept of parameter sharing has been previously studied in various domains, from the core foundation of the design principle of convolutional and recurrent neural networks to parallel sequence models, such as the Transformer \cite{dehghani2018universal, lan2019albert}. The most notable example is \cite{lan2019albert} in the natural language processing domain, which demonstrated a significantly reduced memory footprint of BERT \cite{devlin2018bert} using a cross-layer parameter sharing of the self-attention block. We investigated the effectiveness of the weight-sharing concept on different granularities for flow-based models. We applied parameter sharing on a \textit{model level}, where a shared neural density estimator was applied to multiple stages of bijective transformation that performed bijective operations. Contrary to \cite{lan2019albert}, our study revealed the following findings: in NFs, sharing an entire block failed to competitively model the probability density, whereas minimal relaxation from the decomposition was critical to the performance.
 
 It is noteworthy that continuous-time normalizing flows (CNF) \cite{grathwohl2019ffjord, chen2018neural} features a form of the "shared" neural network $f$. The central difference between CNF and NanoFlow (and non-continuous NFs in general) is that CNF formulates the transformation by an ordinary differential equation (ODE) with an iterative evaluation of $f$ to reach an error tolerance of the adaptive ODE solver, whereas NFs directly model pre-defined steps of transformation with $f_k$ (or $f$ in NanoFlow) with a single network pass. The effectiveness and potential benefits of the shared $f$ outside the ODE-based CNFs are yet to be studied in the literature, which we aim to systematically address with NanoFlow.

\section{Discussion}
In this study, we presented an extensive and systematic analysis of the feasibility and potential benefits of using a single shared neural density estimator for multiple flow operations. Based on the analysis, we developed a novel parameterization scheme called NanoFlow, which enabled scalable NFs with a nearly constant memory complexity and competitive performance as both a generative and a density estimation model, owing to the compact network capacity. This enables direct control over the tradeoff between expressiveness and inference latency, which is beneficial in domains where compact parameterization is desired. The target performance can be explicitly designed using NanoFlow as a building block depending on the task requirements, which can be useful for practitioners who incorporate NFs into applications.

The decomposed view on building flow-based models with NanoFlow suggests that two directions can be endeavored in future research: composing more expressive bijections, which has been the primary focus in existing literature, and building an optimized neural density estimator that can potentially provide a more adaptive computation path leveraged by flow indication embedding. Furthermore, these proposed future studies can be expanded from \cite{ho2019flow++}, which investigated better neural architecture designs for building flow-based models using self-attention for the estimator. Combined with increasing evidence in other research domains applying similar architecture \cite{lan2019albert}, we expect the self-attention-based estimator to provide more expressive density estimations \cite{fakoor2020trade, parmar2018image}, where the attention mechanism could be directly augmented from flow indication embedding. We leave this research direction for future works.

In summary, NanoFlow, which is a bijection-agnostic and generalized solution that achieves significant savings in network capacity, provides an alternative method for parameterizing NFs. Extensive experiments on real-world data domains have provided deep insights into the relationship between the capacity of deep generative models and the expressiveness of flow operations, along with possible future research directions. We hope that the modular scheme of NanoFlow will motivate researchers to further develop flexible and scalable flow-based models.

\section*{Broader Impact}
The main motivation of this study was to observe a major hurdle in incorporating a powerful generative capability of NFs to various application domains, where we need significantly larger neural network capacity to reach the desired level of performance. As our work would impact the practicality of NFs as a mainstream probabilistic toolkit, practitioners should be cautious about possible misrepresentations of our flow indication embedding methods depending on how one further augments the embedding to specific tasks of interest.

In particular, although we demonstrated that our flow indication embedding is domain agnostic and independent variables, it is possible to incorporate task-specific priors into our framework, which can potentially achieve better control of the latent space. By contrast, there is a risk of potential misinterpretation of the embedding, together with the latent space, from biases inside the dataset. Because NFs have exact latent spaces that can be useful for downstream tasks such as facial manipulation \cite{kingma2018glow}, it would have a higher chance of direct exposure to various levels of biases. This could result in a potential exploitation of our embedding methods as an explainable or predictive embedding vector of the biased aspects that could be inherent in the data. Considering these possible directions for the downstream applications of NFs, one should be cautious about extrapolating our embedding scheme in attempts to build improved embedding methods for the target tasks, particularly when leveraging priors into the independent variables we demonstrated.

\section*{Acknowledgements}
We thank Wei Ping for helpful discussion and feedback on implementation details of WaveFlow \cite{ping2020waveflow} model. We also thank Heeseung Kim for careful proofreading. This work was supported by the BK21 FOUR program of the Education and Research Program for Future ICT Pioneers, Seoul National University in 2020 and the National Research Foundation of Korea (NRF) grant funded by the Korea government
(Ministry of Science and ICT) [No. 2018R1A2B3001628].

\bibliography{neurips_2020}

\begin{thebibliography}{10}

\bibitem{chen2018neural}
Ricky~TQ Chen, Yulia Rubanova, Jesse Bettencourt, and David~K Duvenaud.
\newblock Neural ordinary differential equations.
\newblock In {\em Advances in neural information processing systems}, pages
  6571--6583, 2018.

\bibitem{chen2019residual}
Tian~Qi Chen, Jens Behrmann, David~K Duvenaud, and J{\"o}rn-Henrik Jacobsen.
\newblock Residual flows for invertible generative modeling.
\newblock In {\em Advances in Neural Information Processing Systems}, pages
  9913--9923, 2019.

\bibitem{dehghani2018universal}
Mostafa Dehghani, Stephan Gouws, Oriol Vinyals, Jakob Uszkoreit, and {\L}ukasz
  Kaiser.
\newblock Universal transformers.
\newblock {\em arXiv preprint arXiv:1807.03819}, 2018.

\bibitem{devlin2018bert}
Jacob Devlin, Ming-Wei Chang, Kenton Lee, and Kristina Toutanova.
\newblock Bert: Pre-training of deep bidirectional transformers for language
  understanding.
\newblock {\em arXiv preprint arXiv:1810.04805}, 2018.

\bibitem{dinh2014nice}
Laurent Dinh, David Krueger, and Yoshua Bengio.
\newblock Nice: Non-linear independent components estimation.
\newblock {\em arXiv preprint arXiv:1410.8516}, 2014.

\bibitem{dinh2016density}
Laurent Dinh, Jascha Sohl-Dickstein, and Samy Bengio.
\newblock Density estimation using real nvp.
\newblock In {\em International Conference on Learning Representations}, 2017.

\bibitem{durkan2019neural}
Conor Durkan, Artur Bekasov, Iain Murray, and George Papamakarios.
\newblock Neural spline flows.
\newblock In {\em Advances in Neural Information Processing Systems}, pages
  7509--7520, 2019.

\bibitem{fakoor2020trade}
Rasool Fakoor, Pratik Chaudhari, Jonas Mueller, and Alexander~J Smola.
\newblock Trade: Transformers for density estimation.
\newblock {\em arXiv preprint arXiv:2004.02441}, 2020.

\bibitem{finlay2020train}
Chris Finlay, J{\"o}rn-Henrik Jacobsen, Levon Nurbekyan, and Adam~M Oberman.
\newblock How to train your neural ode: the world of jacobian and kinetic
  regularization.
\newblock In {\em International Conference on Machine Learning}, 2020.

\bibitem{goodfellow2014generative}
Ian Goodfellow, Jean Pouget-Abadie, Mehdi Mirza, Bing Xu, David Warde-Farley,
  Sherjil Ozair, Aaron Courville, and Yoshua Bengio.
\newblock Generative adversarial nets.
\newblock In {\em Advances in neural information processing systems}, pages
  2672--2680, 2014.

\bibitem{grathwohl2019ffjord}
Will Grathwohl, Ricky~TQ Chen, Jesse Bettencourt, Ilya Sutskever, and David
  Duvenaud.
\newblock Ffjord: Free-form continuous dynamics for scalable reversible
  generative models.
\newblock In {\em International Conference on Learning Representations}, 2019.

\bibitem{ho2019flow++}
Jonathan Ho, Xi~Chen, Aravind Srinivas, Yan Duan, and Pieter Abbeel.
\newblock Flow++: Improving flow-based generative models with variational
  dequantization and architecture design.
\newblock In {\em International Conference on Machine Learning}, pages
  2722--2730, 2019.

\bibitem{hoogeboom2019emerging}
Emiel Hoogeboom, Rianne Van Den~Berg, and Max Welling.
\newblock Emerging convolutions for generative normalizing flows.
\newblock In {\em International Conference on Machine Learning}, pages
  2771--2780, 2019.

\bibitem{ljspeech17}
Keith Ito.
\newblock The lj speech dataset.
\newblock \url{https://keithito.com/LJ-Speech-Dataset/}, 2017.

\bibitem{kim2019flowavenet}
Sungwon Kim, Sang-Gil Lee, Jongyoon Song, Jaehyeon Kim, and Sungroh Yoon.
\newblock Flowavenet: A generative flow for raw audio.
\newblock In {\em International Conference on Machine Learning}, pages
  3370--3378, 2019.

\bibitem{kingma2014adam}
Diederik~P Kingma and Jimmy Ba.
\newblock Adam: A method for stochastic optimization.
\newblock {\em arXiv preprint arXiv:1412.6980}, 2014.

\bibitem{kingma2016improved}
Diederik~P Kingma, Tim Salimans, Rafal Jozefowicz, Xi~Chen, Ilya Sutskever, and
  Max Welling.
\newblock Improved variational inference with inverse autoregressive flow.
\newblock In {\em Advances in Neural Information Processing Systems}, pages
  4743--4751, 2016.

\bibitem{kingma2013auto}
Diederik~P Kingma and Max Welling.
\newblock Auto-encoding variational bayes.
\newblock {\em International Conference on Learning Representations}, 2013.

\bibitem{kingma2018glow}
Durk~P Kingma and Prafulla Dhariwal.
\newblock Glow: Generative flow with invertible 1x1 convolutions.
\newblock In {\em Advances in neural information processing systems}, pages
  10215--10224, 2018.

\bibitem{lan2019albert}
Zhenzhong Lan, Mingda Chen, Sebastian Goodman, Kevin Gimpel, Piyush Sharma, and
  Radu Soricut.
\newblock Albert: A lite bert for self-supervised learning of language
  representations.
\newblock {\em arXiv preprint arXiv:1909.11942}, 2019.

\bibitem{ma2019macow}
Xuezhe Ma, Xiang Kong, Shanghang Zhang, and Eduard Hovy.
\newblock Macow: Masked convolutional generative flow.
\newblock In {\em Advances in Neural Information Processing Systems}, pages
  5891--5900, 2019.

\bibitem{oord2018parallel}
Aaron Oord, Yazhe Li, Igor Babuschkin, Karen Simonyan, Oriol Vinyals, Koray
  Kavukcuoglu, George Driessche, Edward Lockhart, Luis Cobo, Florian Stimberg,
  et~al.
\newblock Parallel wavenet: Fast high-fidelity speech synthesis.
\newblock In {\em International Conference on Machine Learning}, pages
  3918--3926, 2018.

\bibitem{papamakarios2017masked}
George Papamakarios, Theo Pavlakou, and Iain Murray.
\newblock Masked autoregressive flow for density estimation.
\newblock In {\em Advances in Neural Information Processing Systems}, pages
  2338--2347, 2017.

\bibitem{parmar2018image}
Niki Parmar, Ashish Vaswani, Jakob Uszkoreit, Lukasz Kaiser, Noam Shazeer,
  Alexander Ku, and Dustin Tran.
\newblock Image transformer.
\newblock In {\em International Conference on Machine Learning}, pages
  4055--4064, 2018.

\bibitem{ping2020waveflow}
Wei Ping, Kainan Peng, Kexin Zhao, and Zhao Song.
\newblock Waveflow: A compact flow-based model for raw audio.
\newblock In {\em International Conference on Machine Learning}, 2020.

\bibitem{prenger2019waveglow}
Ryan Prenger, Rafael Valle, and Bryan Catanzaro.
\newblock Waveglow: A flow-based generative network for speech synthesis.
\newblock In {\em ICASSP 2019-2019 IEEE International Conference on Acoustics,
  Speech and Signal Processing (ICASSP)}, pages 3617--3621. IEEE, 2019.

\bibitem{rezende2015variational}
Danilo Rezende and Shakir Mohamed.
\newblock Variational inference with normalizing flows.
\newblock In {\em International Conference on Machine Learning}, pages
  1530--1538, 2015.

\bibitem{song2019mintnet}
Yang Song, Chenlin Meng, and Stefano Ermon.
\newblock Mintnet: Building invertible neural networks with masked
  convolutions.
\newblock In {\em Advances in Neural Information Processing Systems}, pages
  11004--11014, 2019.

\bibitem{van2016wavenet}
A{\"a}ron Van Den~Oord, Sander Dieleman, Heiga Zen, Karen Simonyan, Oriol
  Vinyals, Alex Graves, Nal Kalchbrenner, Andrew~W Senior, and Koray
  Kavukcuoglu.
\newblock Wavenet: A generative model for raw audio.
\newblock In {\em SSW}, page 125, 2016.

\bibitem{van2016pixel}
Aaron Van~Oord, Nal Kalchbrenner, and Koray Kavukcuoglu.
\newblock Pixel recurrent neural networks.
\newblock In {\em International Conference on Machine Learning}, pages
  1747--1756, 2016.

\end{thebibliography}
\bibliographystyle{plain}

\newpage
\appendix
\section*{Appendix}

\section{Implementation details of flow indication embedding}
\label{appendix:a}
In this section, we describe the implementation details of flow indication embedding used in this study, depending on the architecture. We used additive bias and multiplicative gating for WaveFlow-based experiments and concatenative embedding and multiplicative gating for Glow-based experiments.

\textbf{Concatenative embedding.} At the start of each flow, we concatenated the input $\bm{X}$ with $\bm{e}^{k}$ as the augmented representation as follows:
\begin{equation}
\bm{X}_{cat}=Concatenate(\bm{X}, \bm{e}^{k}).
\end{equation}

The $\bm{e}^{k}$ was reshaped to match the shape of the input for each flow. For Glow-based experiments, we reshaped $\bm{e}^{k}$ as $\bm{e}^{k}\in \mathbb{R}^{\hat{C}\times height \times width}$, where $\hat{C}=\frac{D}{height \times width}$, and performed concatenation along the channel-axis.

\textbf{Additive bias.} We used the notation $h^{k, l}\in \mathbb{R}^{H, \cdot}$ as the hidden representation of the $l$-th layer from the $k$-th flow, where $H$ is the number of hidden channels of the neural network. We applied the channel-wise additive bias to $h^{k, l}$ using a single fully connected layer for projection as follows:
\begin{equation}
\tilde{h}^{k, l} = h^{k, l} + W^l\bm{e}^k,
\end{equation}
where $W^l\bm{e}^k \in \mathbb{R}^H$. After training, we can cache the projected bias from the embedding as the final network parameters and discard the projection weights. The reported parameter count in Table \ref{mos} is obtained from the trained model with the projection weights discarded.

\textbf{Multiplicative gating.} We performed multiplicative gating to $h^{k, l}$ by employing a vector $\delta^{k, l} \in \mathbb{R}^{H}$ as follows:
\begin{equation}
\hat{h}^{k, l} = \exp(\delta^{k, l}) \cdot  h^{k, l}.
\end{equation}

$\delta^{k, l}$ was initialized to zero to initially perform the identity. For WaveFlow-based experiments, we applied additive bias followed by multiplicative gating. For Glow-based experiments, we applied multiplicative gating before applying the ReLU activation.

\section{Additional experimental results}
\label{appendix:b}
In this section, we provide additional experimental results.

\textbf{Effect of the number of shared layers.} We measured a tolerance to the decreased network capacity of NanoFlow by varying the number of the shared layers of the network. We used the 8-layer WaveFlow with 64 residual channels as the non-shared baseline, and trained NanoFlow-naive and NanoFlow by partially replacing the layers from the bottom (i.e. closer to the input) with the shared weights. We trained all models with the batch size of two for 600 K iterations under the same learning rate schedule as described in Section \ref{audio_generation_results}. We used $D=128$ for $\bm{e}^k$ of NanoFlow.

\begin{figure}
\centering
    \includegraphics[width=0.6\linewidth]{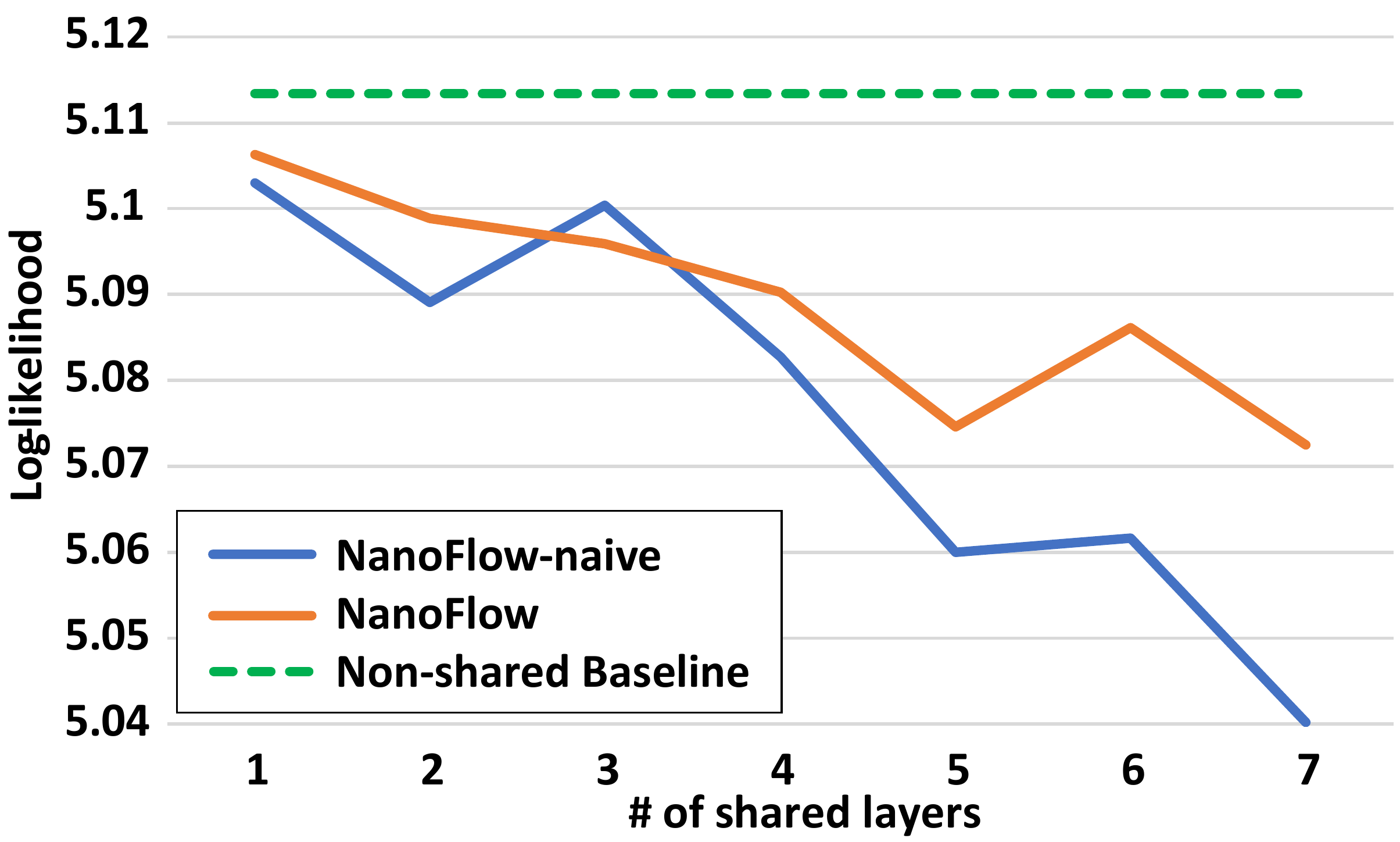}
    \caption{Analysis on the effect of the number of shared layers.}%
    \label{num_layers_shared_result}
\end{figure}

Figure \ref{num_layers_shared_result} shows differences in the performance drop from the varied amount of the decreased network capacity. We can see that NanoFlow-naive degraded its performance more significantly than our final model, which indicates that the proposed technique provided better tolerance to the decreased capacity of the network.

\begin{table}[!htb]
%\tiny\setlength{\tabcolsep}{1pt}
\begin{minipage}{.49\linewidth}
\centering
\caption{Additional results of using non-affine coupling with rational-quadratic splines.}
\label{nsf_result}
\medskip
\begin{tabular}{lcc} \toprule
  Method & Params (M) & LL \\ \midrule
WaveFlow & 22.336 & 5.2059 \\
NanoFlow & 2.819 & 5.1586 \\ \midrule
WaveFlow + RQ-NSF & 22.432 & 5.1866 \\
NanoFlow + RQ-NSF & 2.915 & 5.1614 \\
\bottomrule
  \end{tabular}
\end{minipage}\hfill
\begin{minipage}{0.49\linewidth}
\centering
\caption{Results when applying the method to the reference network topology of Glow model (NanoFlowAlt), evaluated at 600 epochs.}
\label{tab:second_table}
\medskip
\begin{tabular}{lcc} \toprule
Method (600 epochs) & Params (M) & bpd \\ \midrule
Glow (256 channels) & 15.973 & \textbf{3.44} \\
NanoFlowAlt-naive & 0.778 & 3.75 \\
NanoFlowAlt-decomp & 6.783 & 3.54 \\
NanoFlowAlt & 6.961 & 3.53\\
NanoFlowAlt (K=48) & 10.319 & 3.51 \\
\bottomrule
\end{tabular}
\end{minipage} 
\end{table}

\textbf{Compatibility beyond the affine coupling.} Our main experiments used WaveFlow \cite{ping2020waveflow} and Glow \cite{kingma2018glow}, where both models used affine coupling for the transformation. We show that NanoFlow is not restricted to a specific choice of the bijection by replacing the affine coupling for WaveFlow-based models with rational-quadratic splines (RQ-NSF) \cite{durkan2019neural}. We trained both WaveFlow and NanoFlow with the same training strategy as described in Section \ref{audio_generation_results}. We used the the following hyperparameters for the rational-quadratic spline: the number of bins of 32 and the tail bound of 5. We experienced unpleasing popping sounds from the generated audio for all models if we set these values lower.

The likelihood results from Table \ref{nsf_result} show that NanoFlow + RQ-NSF performed slightly better than the default affine coupling, whereas the high-capacity WaveFlow scored worse likelihood. We are not drawing any conclusive claim regarding the different classes of the bijection based on these observations as we have not performed an exhaustive hyperparameter search and training schedule. However, the results indicate that NanoFlow is not restricted to the particular coupling and can be applied to various other classes of flows.

\textbf{Caveats}. As demonstrated in the main experimental results, NanoFlow is designed for leveraging the rich representational power of the deep neural network. In other words, a careful allocation of the parameters is required under NanoFlow framework, where the shared estimator should have sufficient capacity, while keeping the non-shared projection layers lightweight.

We additionally show a negative result when the aforementioned caveats are not met. We trained the NanoFlow variants of Glow with the exact same network topology: from $3\times3$ conv $\rightarrow$ $1\times1$ conv $\rightarrow$ $3\times3$ projection conv layers per flow, NanoFlowAlt shared the first two layers and used the separate $3\times3$ projection conv. Results showed that the model performed significantly worse than the baseline architecture, even though NanoFlowAlt (K=48) has similar network size (10 M) to our main result. This indicates that we have to assure that the shared neural density estimator possesses the sufficient capacity.

\newpage
\section{Samples generated from image models}
\label{appendix:c}
\begin{figure*}[h]
        \centering
        \begin{subfigure}[b]{0.475\textwidth}
            \centering
            \includegraphics[width=\textwidth]{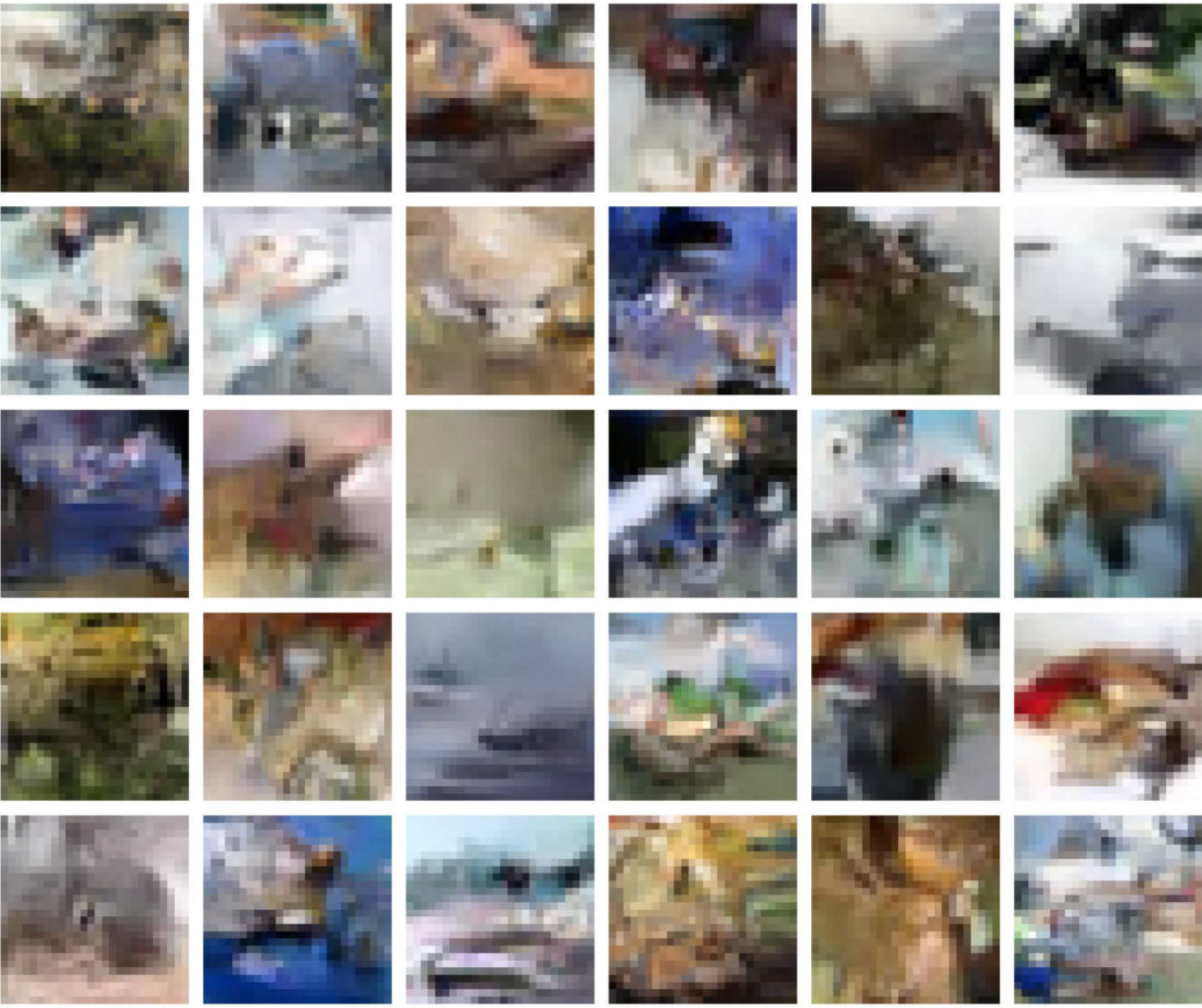}
            \caption[Glow]%
            {{\small Glow (bpd = 3.40)}}    
            \label{figa:1}
        \end{subfigure}
        \hfill
        \begin{subfigure}[b]{0.475\textwidth}  
            \centering 
            \includegraphics[width=\textwidth]{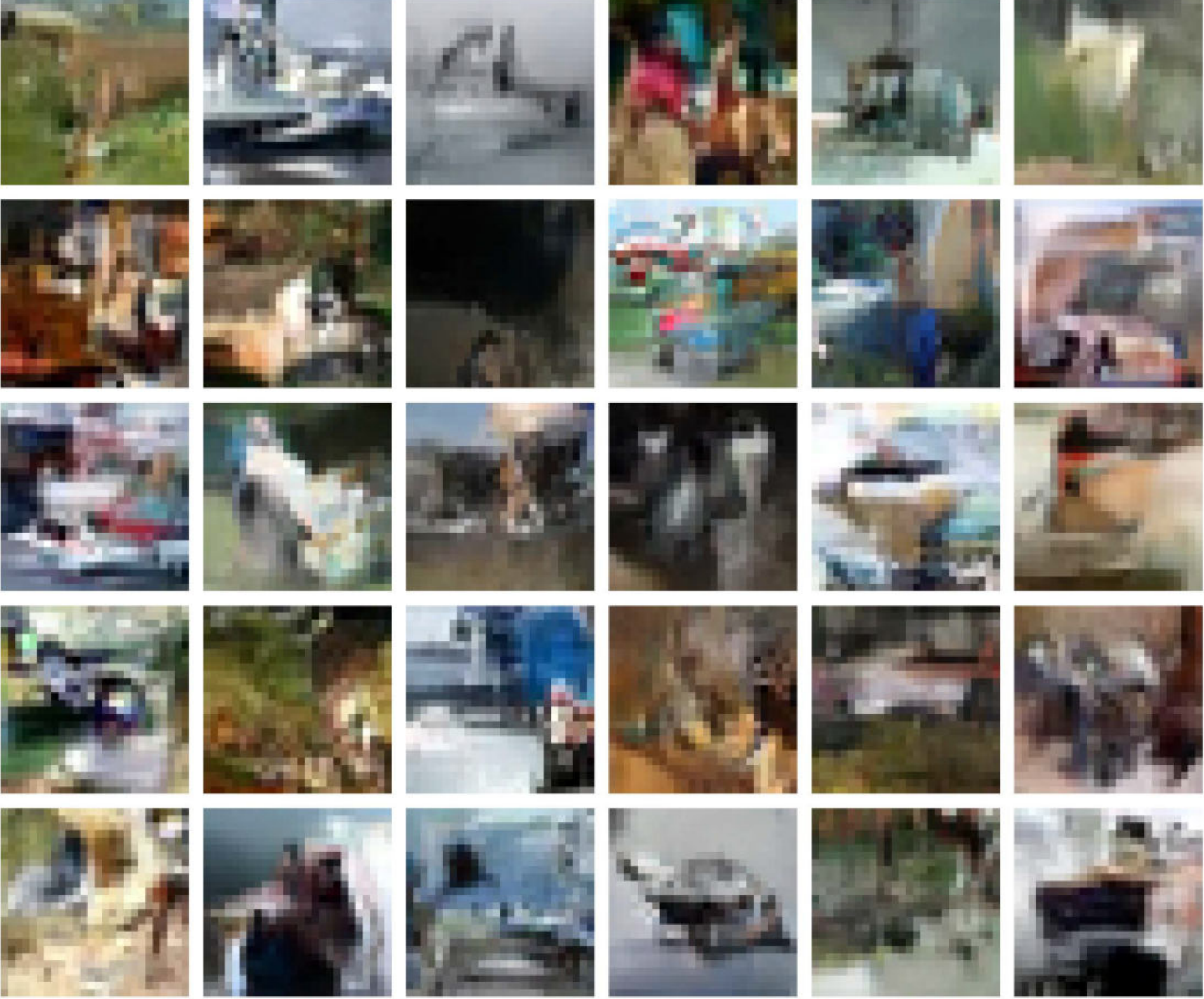}
            \caption[Glow-large]%
            {{\small Glow-large (bpd = 3.30)}}    
            \label{figa:2}
        \end{subfigure}
        \vskip\baselineskip
        \begin{subfigure}[b]{0.475\textwidth}   
            \centering 
            \includegraphics[width=\textwidth]{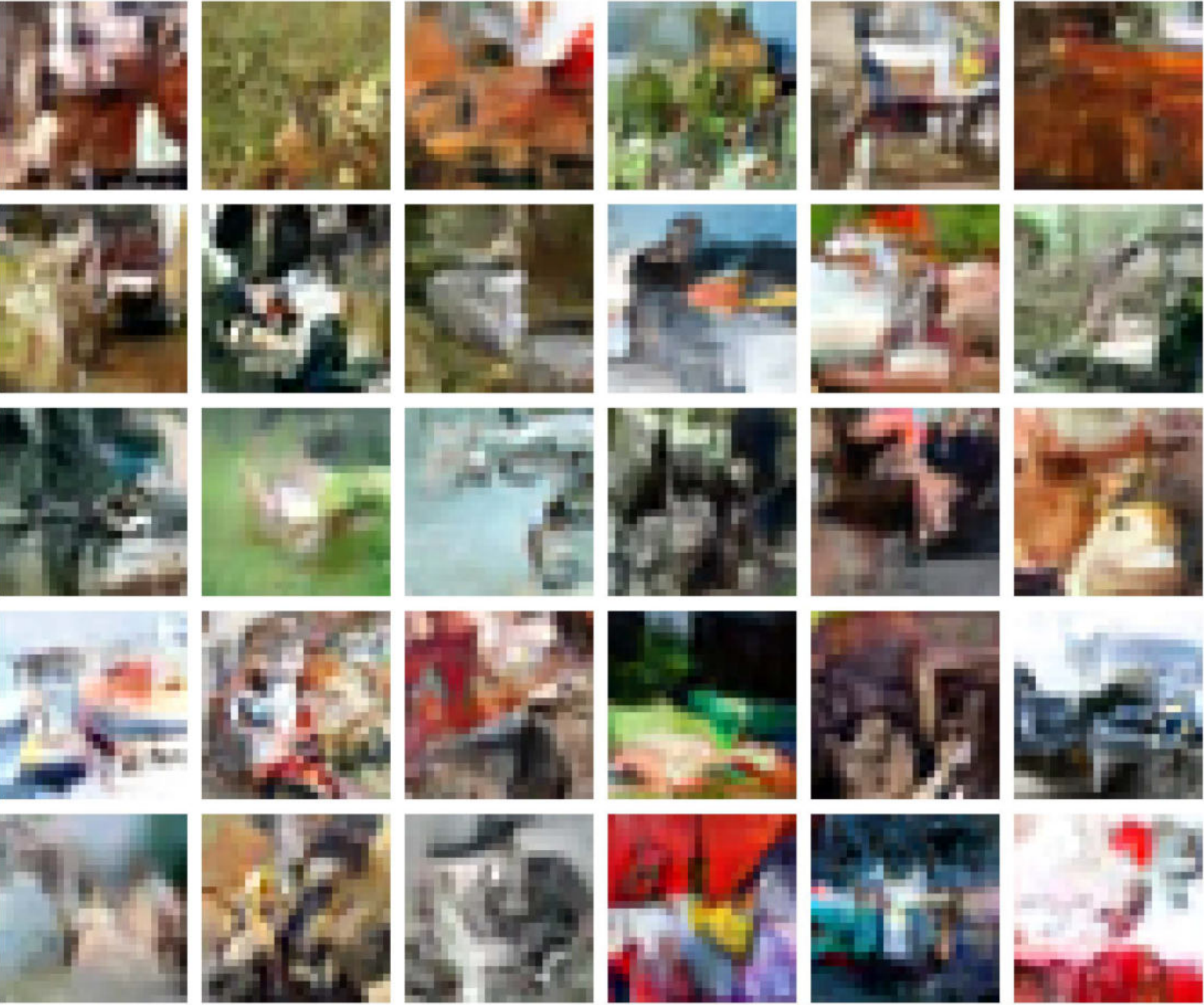}
            \caption[NanoFlow-naive]%
            {{\small NanoFlow-naive (bpd = 3.40)}}    
            \label{figa:3}
        \end{subfigure}
        \hfill
        \begin{subfigure}[b]{0.475\textwidth}   
            \centering 
            \includegraphics[width=\textwidth]{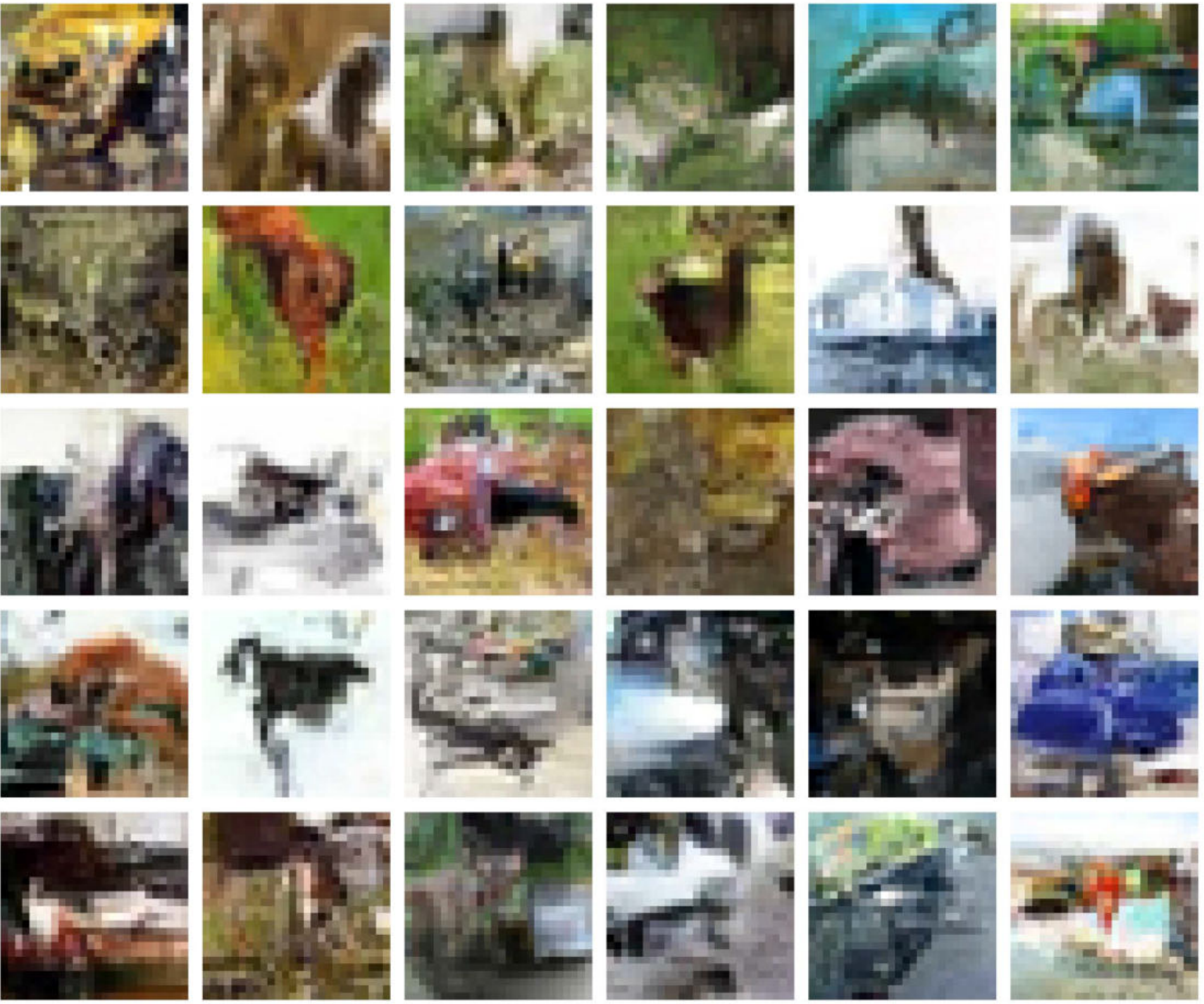}
            \caption[]%
            {{\small NanoFlow (K = 48) (bpd = 3.25)}}    
            \label{figa:4}
        \end{subfigure}
        \caption[Sample]
        {\small Unconditional samples generated from image models in Section \ref{image_density_modeling} trained on CIFAR10. The temperature was set to 1.0. Models with lower bpd tended to generate sharper and detailed textures, which is consistent with the existing literature.} 
        \label{figa}
    \end{figure*}

\end{document}